\documentclass{article} 
\usepackage{iclr2025_conference,times}
\usepackage{booktabs} 
\usepackage{graphicx}
\usepackage{multirow} 
\usepackage{wrapfig}

\usepackage{amsmath,amsfonts,bm}

\def\eqref#1{equation~\ref{#1}}

\def\1{\bm{1}}

\DeclareMathAlphabet{\mathsfit}{\encodingdefault}{\sfdefault}{m}{sl}
\SetMathAlphabet{\mathsfit}{bold}{\encodingdefault}{\sfdefault}{bx}{n}

\usepackage{hyperref}
\usepackage{url}

\title{\textit{PEAR}: \textit{P}osition-\textit{E}mbedding-\textit{A}gnostic Attention \textit{R}e-weighting Enhances Retrieval-Augmented Generation with Zero Inference Overhead}

\author{Tao Tan$^{1}$\thanks{Equal Contribution.} \quad Yining Qian$^{2 *}$\quad Ang Lv$^{1 *}$\quad Hongzhan Lin$^1$\quad Songhao Wu$^1$\\
\textbf{Yongbo Wang}$^3$\quad \textbf{Feng Wang}$^3$\quad \textbf{Jingtong Wu}$^{3}$\quad \textbf{Xin Lu}$^3$\quad \textbf{Rui Yan}$^{1}$\thanks{Corresponding author: Rui Yan (ruiyan@ruc.edu.cn).}\\
$^{1}$Gaoling School of Artificial Intelligence, Renmin University of China\\
$^2$Southeast University\\
$^3$Ant Group\\
\\
\url{https://github.com/TTArch/PEAR-RAG}
}
\iclrfinalcopy 
\begin{document}

\maketitle

\begin{abstract}
Large language models (LLMs) enhanced with retrieval-augmented generation (RAG) have introduced a new paradigm for web search. However, the limited context awareness of LLMs degrades their performance on RAG tasks. 
Existing methods to enhance context awareness are often inefficient, incurring time or memory overhead during inference, and many are tailored to specific position embeddings. 
In this paper, we propose \textbf{P}osition-\textbf{E}mbedding-\textbf{A}gnostic attention \textbf{R}e-weighting (\textit{PEAR}), which enhances the context awareness of LLMs with zero inference overhead.
Specifically, on a proxy task focused on context copying, we first detect heads which suppress the models' context awareness, thereby diminishing RAG performance.
To weaken the impact of these heads, we re-weight their outputs with learnable coefficients.
The LLM (with frozen parameters) is optimized by adjusting these coefficients to minimize loss on the proxy task.
During inference, the optimized coefficients are fixed to re-weight these heads, regardless of the specific task at hand.
Our proposed \textit{PEAR} offers two major advantages over previous approaches: 
(1) It introduces zero additional inference overhead in terms of memory usage or inference time, while outperforming competitive baselines in accuracy and efficiency across various RAG tasks.
(2) It is independent of position embedding algorithms, ensuring broader applicability.
\end{abstract}

\section{Introduction}
Retrieval-augmented generation (RAG,~\citep{lewis2021retrievalaugmentedgenerationknowledgeintensivenlp}) is widely utilized to enhance large language models (LLMs) on tasks like question answering.
Typically, an RAG framework retrieves documents related to users' questions from external knowledge bases or web pages, and then arranges them in the LLMs' context as the references to form answers.
This LLM-based question-answering paradigm has given rise to a promising web search paradigm~\citep{new_bing,openai2024gpt4technicalreport}.

Recent research demonstrated LLMs' limitations on context awareness, especially when processing long context.
These limitations in LLMs' context awareness challenge the effectiveness and robustness of RAG frameworks.
For instance, Liu et al.~\citep{liu2023lost} found that when performing in-context retrieval tasks, LLMs exhibit insensitivity to information located in the middle of the context, a phenomenon referred to as ``lost in the middle.''
Chen et al.~\citep{chen-etal-2024-fortify} identified a mathematical property in rotary position embedding (RoPE,~\citep{su2023roformerenhancedtransformerrotary}) that results in LLMs assigning less attention to specific contextual positions, leading to varying context awareness throughout the entire context.

Existing approaches to enhancing LLMs' context awareness are inefficient in terms of memory and time cost.
Some works~\citep{peysakhovich2023attention} segment and re-arrange the input context, with the assumption that placing important information in positions the model attends well can improve RAG's effectiveness. 
This method incurs additional inference time costs, negatively affecting user experience, as it requires multiple forward passes to obtain attention weights for guiding segment rearrangement.
Another group of studies modifies the model's working mechanism, specifically by employing a set of position embeddings to adjust the attention preferences of attention heads.
While these methods are input-agnostic, they also lack efficiency due to ``parallelable'' forward passes~\citep{chen-etal-2024-fortify} with slightly increased time cost in additional aggregation operation, or disrupt the parallelism of multi-head attention (resulting in increased time cost) to achieve low memory cost~\citep{lin2024mixtureincontextexpertsenhance}, alternatively, still requiring ``non-parallelable'' multiple forward passes~\citep{zhang2024middlelanguagemodelsuse} which is similar to~\citep{peysakhovich2023attention}. 
Moreover, these studies are mainly designed for RoPE and face challenges in generalizing to other position embedding algorithms, limiting their broader applications.

In this paper, we introduce \textbf{P}osition-\textbf{E}mbedding-\textbf{A}gnostic attention head \textbf{R}e-weighting (PEAR), which unleashes the context awareness of LLMs, thereby improving their RAG performance.
\textit{PEAR} achieves zero additional overhead in memory usage and inference time.
Our motivation relies on the following facts:

\begin{enumerate}
    \item Prior research~\citep{mcdougall2023copysuppressioncomprehensivelyunderstanding,lv2024interpretingkeymechanismsfactual} detected some attention heads decreasing the language model's prediction confidence by suppressing the flow of contextual information to the final position within the context, where the output is to be generated.
    \item This suppression negatively impacts LLMs' context awareness, particularly abilities in in-context retrieval and context integration, which are crucial for effective RAG.
\end{enumerate}

As a result, we contend that such suppression mechanism in LLMs can be safely\footnote{``Safely'' means that the parametric knowledge and fundamental capabilities remain unaffected.
Detailed experimental results are presented in Section~\ref{sec:safe}.} weakened in RAG scenarios, thereby improving the RAG performance of LLMs. 
Our proposed \textit{PEAR} includes two stages:

\begin{table}[t]
    \centering
    \caption{An example input for the proxy task, where unique letters representing distinct tokens, with \( n = 4 \). 
    For example, at position \( n + i = 5 \) (with \( i = 1 \)), when an LLM receives ``ABCDA'' as input, it is likely to output ``B.'' 
    This happens because the last occurrence of ``A'' in the preceding context is followed by ``B.''
    If a head suppresses copying ``B'' from position \( i + 1 = 2 \) to position \( n + i = 5 \), it could negatively impact RAG performance.}
    \label{tab:input_demo}
    \vspace{5pt}
    \begin{tabular}{l|cccccccc}\toprule
       Input Sequence  & A & B & C & D & A & B & C & D \\\hline
       \noalign{\vskip 1mm}
       Position Index  & 1 & 2 & 3 & 4 & 5 & 6 & 7 & 8\\\bottomrule
    \end{tabular}
\end{table}

In the first stage, we discover attention heads negatively affect performance on a proxy task. 
The proxy task involves feeding the model a random token sequence of length $2n$, which consists of a duplicated sub-sequence of length $n$. Table~\ref{tab:input_demo} illustrates an example input. 
At position $n+i$, the model typically predicts the token from position $i+1$, as the natural continuation for a semantically meaningless context is to copy the existing in-context token pattern~\citep{lv2024languagemodelsgrokcopy}. 
The negatively impactful attention heads are discovered using the path patching technique~\citep{wang2023interpretability}.
Since this proxy task is free from semantic bias and requires both in-context retrieval and generation based on the context\textemdash fundamental capacities for RAG, we refer to discovered heads as RAG-suppression heads\footnote{We do not imply that these heads hinder RAG through the same mechanisms (discussed in Section~\ref{head_detection}).}. 

In the second stage, we weaken detected RAG-suppression heads by re-weighting their outputs using learnable coefficients. 
These coefficients are optimized by minimizing the LLM's loss on the proxy task, with the objective of next-token prediction in a supervised fine-tuning process (loss is computed only for the second half of the random-token sequence). 
During the optimization, the original LLM parameters are frozen.
Intuitively, most of the learned coefficients are optimized to values less than one, reducing the relative weight of these heads compared to others in the same layer when multi-head outputs are aggregated.
Consequently, their influence during the forward pass is weakened. 
Once optimized, the coefficients remain fixed and are agnostic to downstream RAG tasks.

\textit{PEAR} achieves two-fold contributions:

\begin{enumerate}  
    \item \textit{PEAR} introduces zero inference overhead in terms of both memory usage and inference time—an advantage not achieved by competitive baselines. 
    Across a wide range of RAG tasks, \textit{PEAR} surpasses previous works in both efficiency and accuracy.
    \item \textit{PEAR} is independent of specific position embedding algorithms, making it broadly applicable. 
    We demonstrate that \textit{PEAR} enhances the RAG performance of various LLMs using distinct position embeddings (e.g., learnable embeddings~\citep{zhang2022optopenpretrainedtransformer}, RoPE~\citep{su2023roformerenhancedtransformerrotary}, and Alibi~\citep{press2022trainshorttestlong}).
\end{enumerate}

\section{Related works}

In this section, we discuss two research areas closely related to this paper: enhancements to LLMs' context awareness and studies on mechanistic interpretability.

\subsection{Context awareness enhancement}
\label{sec:related-context}
Many studies highlighted limitations in LLMs' context awareness. 
For example, Lu et al.~\citep{lu-etal-2022-fantastically} found that the order of in-context learning (ICL) demonstrations significantly affects ICL accuracy. 
Liu et al.~\citep{liu2023lost} demonstrated that LLMs exhibit stronger awareness of content at the beginning and end of context but weaker awareness in the middle, a phenomenon termed ``lost in the middle.''
Chen et al.~\citep{chen-etal-2024-fortify} proposed that LLMs’ context awareness fluctuates across token positions due to mathematical properties in rotary position embedding. 
These challenges impact applications like RAG that rely on robust context awareness.

Several approaches have been proposed to tackle these issues.
However, existing methods often come at the cost of increased inference time or memory overhead.
Attention Buckets (AB,~\citep{chen-etal-2024-fortify}) enhances context awareness by integrating positional information from a set of various RoPE angles, but it incurs significant inference overhead due to multiple parallel forward passes, leading to increased memory usage. 
Ms-PoE~\citep{zhang2024middlelanguagemodelsuse} calculates distinct re-scaling factors for each attention head, requiring multiple non-parallel forward passes that introduce noticeable time delays. MoICE~\citep{lin2024mixtureincontextexpertsenhance} builds on Attention Buckets by employing a Mixture-of-Experts (MoE,~\citep{moe}) approach, treating each RoPE embedding as a unique in-context expert, thereby limiting extra attention computations to within each layer rather than across the entire forward pass.

Our proposed method, \textit{PEAR}, enhances context awareness by weakening the RAG-suppression heads during the forward pass. It introduces no additional modules or extra forward passes, resulting in \textit{zero} additional overhead in memory usage and inference time. 
Additionally, \textit{PEAR} operates independently of position embedding algorithms, making it applicable to more LLMs compared to existing approaches.

\subsection{Mechanistic interpretability}

Investigating the role of a specific head during a forward pass is one of key focuses in mechanistic interpretability research. 
Wang et al.~\citep{wang2023interpretability} reported that in GPT-2 small~\citep{radford2019language}, the 7th head in the 10th layer, termed the ``negative head,'' significantly hinders answer copying from context. 
McDougall et al.~\citep{mcdougall2023copysuppressioncomprehensivelyunderstanding} comprehensively studied this head, suggesting it functions as a self-repair mechanism to prevent overconfident outputs. 
Lv et al.~\citep{lv2024interpretingkeymechanismsfactual} found that negative heads exist across various LLMs, employing different mechanisms to mitigate overconfidence, such as generating counteracting vectors or introducing high-frequency tokens' information. 
This paper does not examine what specific mechanisms the heads employ to suppress RAG performance but instead aims to discover and suppress heads negatively impactful across general RAG tasks. 

Yu et al.~\citep{yu-etal-2023-characterizing} detected two types of heads in Transformer-based language models during counter-factual task execution (where counter-factual knowledge is provided in the context): memory heads, which prefer to use stored knowledge, and in-context heads, which prefer to use facts in the context. 
However, when re-weighting these heads, only reducing the weight of memory heads successfully enhances the model's preference of using contextual knowledge, while enhancing or mitigating in-context heads does not bring much influence.
Moreover, existing results only demonstrate effectiveness of these heads in the ``country-capital'' task (i.e., prompting the model to answer the capital city of the given country). 
Additionally, there is no evidence suggesting that re-weighting these heads improves the comprehensive context awareness of LLMs.
We owe these failures to their head detection method, and the same re-weighting value applied for all the heads of the same type.
By contrast, in our proposed \textit{PEAR}, each individual head is re-weighted by a specific learnable coefficient, which is optimized through a proxy task independent of downstream tasks.

\section{Preliminaries: Discovery of influential attention heads}
\label{sec:preliminaries}

\begin{wrapfigure}{r}{0.45\textwidth}
  \begin{center}
    \includegraphics[width=\linewidth]{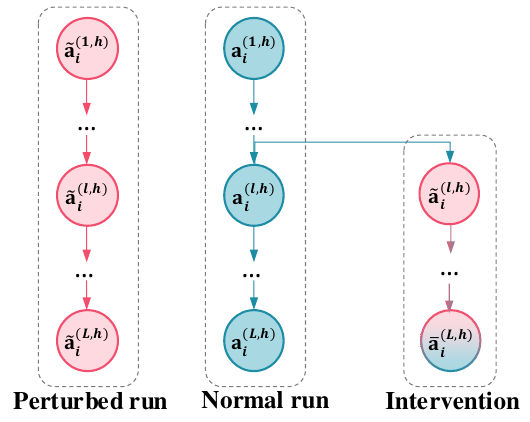}
  \end{center}
  \caption{An illustration of causal mediation methods for circuit discovery.}
  \label{fig:pre}
\end{wrapfigure}

%

For a particular task, research has shown that only a sparse sub-network is activated during the forward pass in Transformer language models~\citep{wang2023interpretability,merullo2024circuit,gong2024mixtureofmodulesreinventingtransformersdynamic}. 
Such a sub-network is referred to as a circuit \citep{olah2020zoom}. 
Discovering circuits provides interpretability into the working mechanisms of language models and offers insights for model enhancement.

The primary method for circuit discovery is based on causal mediation analysis. 
The core idea is to view the forward computation graph as a causal graph, where the output of one module serves as the input for the next. 
In such a case, if the output of a module is changed, the computation of subsequent modules in the causal graph is also affected, as their inputs change.

In this paper, we primarily focus on analyzing the working mechanism of attention heads in language models. 
We briefly introduce a paradigm from a series of works \citep{wang2023interpretability,zhang2024towards,wang2024grokkedtransformersimplicitreasoners} that discovers which attention heads are crucial for processing an input sequence \(X\) of length \(n\). 
Suppose the language model consists of \(L\) layers, with \(H\) attention heads per layer. 
Let \(A^{(l,h)}\) denote the \(h\)-th attention head in the \(l\)-th layer, and let its outputs be denoted by \(\textbf{a}^{(l,h)} \in \mathbb{R}^{n \times d}\). 
We use \(\textbf{a}^{(l,h)}_{i} \in \mathbb{R}^{d}\), where \(1 \le i \le n\), to represent the output at position \(i\).
The discovery paradigm typically includes three steps, as illustrated in Figure~\ref{fig:pre}:

\begin{enumerate}
    \item In the \textit{normal run}, with an input sequence \(X\) (e.g., $X=$``The capital of France is''), \(\textbf{a}^{(l,h)}\) for every attention head are recorded.
    \item In the \textit{perturbed run}, the forward computation runs using the same input sequence \(X\), but with some mediation. 
    This mediation either changes the discrete input tokens within \(X\) by substituting specific keywords (e.g., replacing ``France'' with ``England''), or corrupts the hidden states by adding noise. 
    The modified \(\Tilde{\textbf{a}}^{(l,h)}\) for each attention head are then recorded.
    \item We conduct an intervention on a particular head \( A^{(l,h)} \) at a specific position \( i \) (e.g., the country token position in above examples) in the normal run by substituting its outputs with \(\Tilde{\textbf{a}}^{(l,h)}_{i}\). The subsequent activations in the computational graph are then recomputed (these recomputed activations are denoted as $\bar{\textbf{a}}$ in the figure). 
    If the language model's final output matches the intervention's expectation (e.g., the predicted token changes from "Paris" to "London"), the head $A^{(l,h)}$ is considered to have a positive influence on the processing of sequence $X$. 
\end{enumerate}

This overview outlines a simplified discovery paradigm; detailed measurements of intervention impact are tailored to specific experimental needs.


\section{Methodology}
\label{method}
In this section, we provide a detailed introduction to our proposed method, \textit{PEAR}, which is executed in two stages: (1) discovering RAG-suppression heads and (2) re-weighting coefficient learning.
The first stage discovers attention heads that have a negative impact on general RAG tasks based on circuit discovery for a proxy task.
In the second stage, we optimize learnable coefficients to re-weight the outputs of the discovered heads, aiming to mitigate their RAG-suppression effect.
These coefficients remain fixed during inference, irrespective of the specific input.
Figure~\ref{fig:logit-difference} demonstrates the overview of \textit{PEAR}.

\subsection{Discovery of RAG-suppression heads}
\label{head_detection}

We set up a proxy task and use this task as input for circuit discovery algorithms to discover influential attention heads that hamper LLMs' performance on general RAG tasks.

\paragraph{\textbf{Task Input}} For each input sample, we create a sequence of length \( n \), denoted as \( \{ x_1, \ldots, x_{n} \} \), where each \( x_i \) is a randomly sampled token from the vocabulary. This sequence is repeated to form an input sample \( X = \{ x_1, \ldots, x_{2n} \} \), with \( x_{i} = x_{i+n} \) for \( i \in [1, n] \). 
Research has shown that, in semantically meaningless contexts, models tend to check if the last few tokens in the sequence appeared previously and copy the suffix of their last appearance as the output~\citep{olsson2022context,lv2024languagemodelsgrokcopy}.
We consider an arbitrary LLM to successfully perform the proxy task when, at position \( n+i \), the token with the highest output logits is \( x_{i+1} \).
Table~\ref{tab:input_demo} shows an example input.

This proxy task exhibits two key characteristics that facilitate the effective discovery of RAG-related heads:

\begin{enumerate}
    \item Completing the proxy task requires LLM capabilities essential for a robust RAG framework, such as in-context retrieval and generation based on context, making it suitable for discovering RAG-related attention heads.
    \item The random token composition in \( X \) ensures semantically meaningless input, minimizing knowledge bias and thereby ensuring the discovered attention heads to have general RAG-related functions, independent of specific downstream tasks.
\end{enumerate}

\paragraph{Head discovery}

We previously outlined the head discovery algorithm in Section~\ref{sec:preliminaries}. Here, we provide additional practical details for the first stage of \textit{PEAR}.

\begin{enumerate}
    \item During the \textit{normal run}, the input sequences \(X\) are constructed as above described, with a length of \(2n\).
    \item In the \textit{perturbed run}, we do not modify the input or hidden states; instead, we average the outputs of each attention head along the sequence dimension and record the resulting mean vectors.
    \item We focus on detecting changes in logits at position \(2n-1\), where the model is expected to copy the token from position \(n\). Consequently, we intervene at \(\textbf{a}^{(l,h)}_{2n-1}\) by replacing it with the saved mean vectors.
    \item Our intervention measurements are based on the logits difference, defined as:
    
    \begin{equation}
        \Delta\pi^{(l,h)} = \frac{\Tilde{\pi}^{(l,h)}_{2n-1}[x_{n}]}{\pi_{2n-1}[x_{n}]} - 1,
    \end{equation}
    
    where \(\pi_{2n-1}\) represents the final logits at position \(2n-1\) during the \textit{normal run}, and \([x_{n}]\) denotes selecting the value of the token \(x_{n}\) from the logits. 
    \(\Tilde{\pi}^{(l,h)}\) indicates the logits after intervention on \(A^{(l,h)}\).
    We contend that a higher value of this metric suggests a stronger suppression effect from \(A^{(l,h)}\).
    \item For an arbitrary LLM, we repeat the proxy task multiple times with varying values of $n$ to mitigate bias in context length. 
    The final metric score for each head is the average of the results from these repeated experiments.
    The detailed setup is provided in Section~\ref{sec:exp-setup}.
    \item Based on the final metric scores, we identify the heads with the top-$K$ most negative influence on the proxy task as a set $\mathcal{S}$, defined as:
    \[\mathcal{S} = \{A^{(l,h)} | A^{(l,h)} \text{ has one of the top-$K$ values of } \Delta\pi^{(l,h)}\}.\] 
    These heads are collectively referred to as RAG-suppression heads. 
\end{enumerate}

Notably, we do not suggest that the heads we discovered suppressing RAG tasks operate through the same mechanisms. 
In Section~\ref{sec:tau}, we demonstrate that these heads may serve various functions, such as copy suppression~\citep{mcdougall2023copysuppressioncomprehensivelyunderstanding}, incorporating high-frequency token information~\citep{lv2024interpretingkeymechanismsfactual}, or influencing the behavior of other heads to indirectly affect outputs~\citep{wang2023interpretability}. 
Analyzing the specific mechanism for each head is not the focus of this paper and is left for future works.

\subsection{Re-weighting coefficient learning}
\label{re-weighting_learning}
\begin{wrapfigure}{r}{0.34\textwidth}
\vspace{-4mm}
    \centering
    \includegraphics[width=\linewidth]{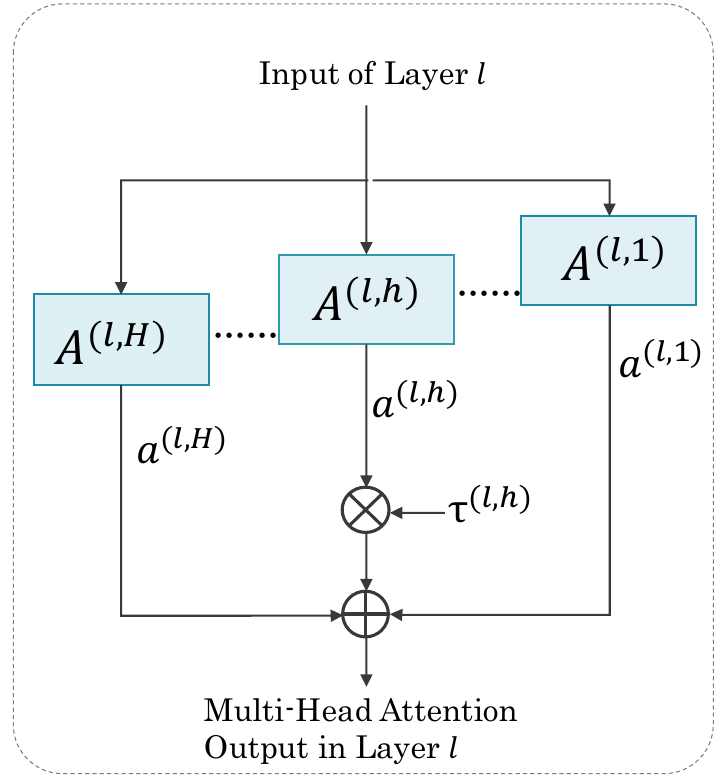}
    \caption{Suppose in layer \( l \), \( A^{(l,h)} \) is discovered as a RAG-suppression head. \textit{PEAR} re-weights its output with a learnable scalar \( \tau^{(l,h)} \).}
    \label{fig:logit-difference}
\end{wrapfigure}
\paragraph{\textbf{Optimization}}
In standard multi-head attention mechanisms, the outputs of all attention heads are aggregated with equal weighting. 
We propose that re-weighting these relative aggregation weights to values less than 1 can mitigate the RAG-suppression effect from our discovered heads. 
To implement this, we modify the forward computation by multiplying the output of each head, $A^{(l,h)}$, in the set $\mathcal{S}$ by a learnable scalar, $\tau^{(l,h)}$, referred to as the re-weighting coefficient. 
The modified output for each head is: 

\begin{equation} 
\textbf{a}^{(l,h)} = \tau ^ {(l,h)} * \textbf{a}^{(l,h)}, \ \text{for each } A^{(l,h)} \in \mathcal{S}. 
\label{eq:2}
\end{equation}

To optimize these re-weighting coefficients for RAG-suppression heads, we freeze the original parameters of the LLM and train only the re-weighting coefficients to minimize the loss on a proxy task. 
Importantly, the loss is calculated only over the latter half of the sequence, optimizing the coefficients to enhance in-context retrieval capacities rather than predicting the next token.
Formally, our adopted loss can be written as:

\begin{equation} 
\mathcal{L} = -\sum^{2n-1}_{i=n} \log p(x_{i+1}|x_{1:i}) 
\end{equation}

Figure~\ref{fig:logit-difference} illustrates a re-weighting process during optimization. 
Notably, the re-weighting process shown in this figure adds extra multiplication operations in a forward pass. 
In practice, when coefficient learning ends, we re-scale $W^{(l,h)}_O$ (the output projection matrix in head $A^{(l,h)}$) by $\tau^{(l,h)}$, which is equivalent to Eq.~\ref{eq:2} and does not add any extra computation during inference.

\paragraph{Inference on Downstream Tasks} 
We highlight several points regarding the inference process of our proposed \textit{PEAR} on downstream RAG tasks:

\begin{enumerate} 
\item In downstream RAG tasks, the re-weighting coefficients are task-independent and remain fixed. 
\item RAG-suppression heads are optimized once for each LLM via the proxy task. 
For a new RAG task, head discovery and coefficient learning do not need to be repeated. 
\end{enumerate}

In theory, our approach, \textit{PEAR}, introduces \textit{zero} additional overhead during inference on downstream RAG tasks, as it does not incorporate extra computational modules; instead, it only adjusts the aggregation weights of specific heads. 
Additionally, the learning of re-weighting coefficients is independent of the LLM architecture, thus making our method compatible with various position embedding algorithms.

\section{Experiments}
\label{sec:exp}
\subsection{Setup}
\label{sec:exp-setup}

In this section, we introduce the LLMs we used for experiments, the baseline methods for enhancing context awareness, the setups for the proxy tasks, hyperparameters for learning re-weighting coefficients.

\paragraph{\textbf{Models and baselines}}

We conducted experiments with three LLMs, each employing a different position embedding algorithm: Llama2-7B-chat-4k~\citep{touvron2023llama2openfoundation} using RoPE~\citep{su2023roformerenhancedtransformerrotary}, OPT-6.7B-2k~\citep{zhang2022optopenpretrainedtransformer} using learnable position embeddings, and Baichuan-13B-4k~\citep{baichuan} using Alibi position embeddings~\citep{press2022trainshorttestlong}.

We also compared several competitive baseline methods for enhancing LLMs' context awareness, including Attention Buckets (AB,~\citep{chen-etal-2024-fortify}), Ms-PoE~\citep{zhang2024middlelanguagemodelsuse}, and MoICE~\citep{lin2024mixtureincontextexpertsenhance}. 
Details on these methods can be found in Section~\ref{sec:related-context}.

\begin{table}[t]
    \centering
    \caption{Discovered RAG-suppression heads in Llama2-Chat-7B-4k, OPT-6.7B-2k and Baichuan-13B-chat-4k, respectively.}
    \label{tab:rag-head}
    \vspace{5pt}
    \resizebox{0.98\linewidth}{!}{
    \begin{tabular}{l|cc}
    \toprule
    Model Name & $(l,h)$ for discovered $A^{(l,h)}$\\\hline
        & (26, 28), (11, 6), (14, 15), (30, 9), (18, 9), (15, 10), (13, 9), (12, 10), (15, 14), (10, 18),  \\
     Llama2-7B-chat-4k   & (15, 25), (19, 15), (29, 15), (14, 0), (10, 2),
        (31, 17), (8, 22), (17, 0), (20, 26), (9, 13), \\
        & (13, 14), (7, 9), (10, 1), (15, 12), (11, 9), (15, 7), (9, 16), (26, 9), (28, 22), (15, 2) \\
        \hline
        \multirow{2}{*}{OPT-6.7B-2k} &  (29, 19), (0, 22), (0, 6), (26, 16), (26, 15), (30, 19), (0, 18), (23, 30), (0, 10), (31, 31),\\
        & (28, 6), (30, 30), (21, 27), (0, 17), (31, 25), (12, 23), (22, 16), (0, 0), (23, 0), (0, 1), (24, 31), (23, 8)\\
         \hline
        \multirow{2}{*}{Baichuan-13B-chat-4k} & (26, 22), (33, 25), (28, 26), (32, 13), (23, 20), (25, 24), (19, 20), (38, 16), (22, 21), (21, 12), \\
        & (3, 24), (39, 39), (20, 27), (37, 21), (0, 32), (24, 39), (39, 28), (39, 20), (27, 24), (2, 20), (36, 10) \\
    \bottomrule
    \end{tabular}}
\end{table}

\begin{figure}
\vspace{-2mm}
    \centering
    \includegraphics[width=0.98\linewidth]{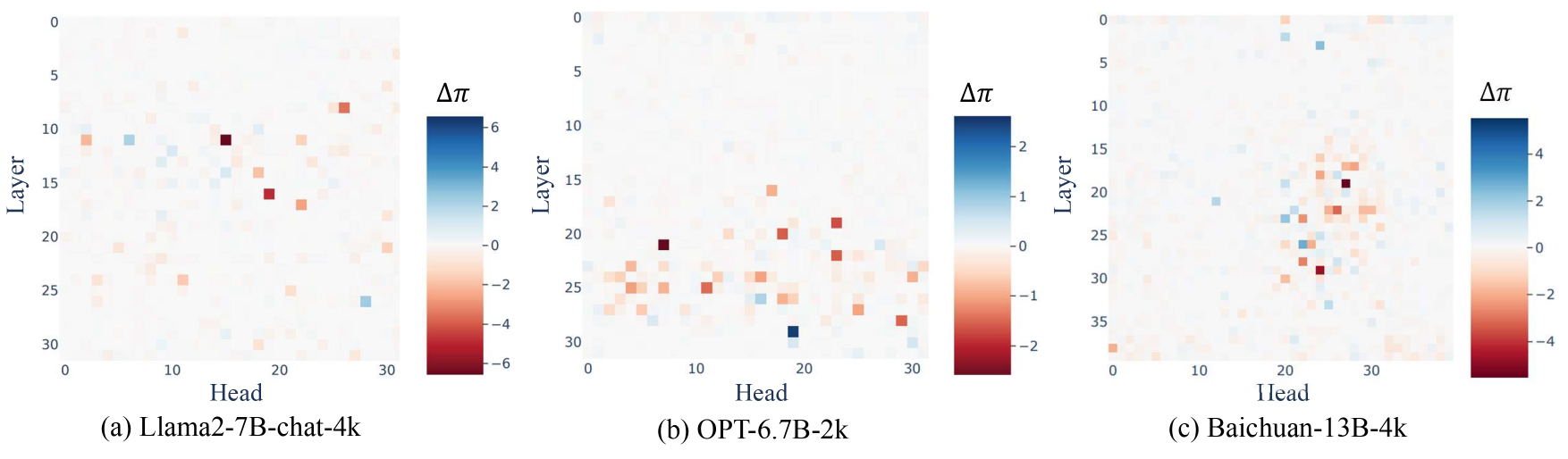}
    \caption{Heatmaps of $\Delta\pi$ scores for each head across three LLMs ($n=10$).}
    \label{fig:heads}
    \vspace{-2mm}
\end{figure}

\begin{table}[t]
\centering
\caption{Performance comparison of Llama2-7B-chat-4k and its enhancements across three RAG tasks.}
\label{table:lb}
\vspace{5pt}
\renewcommand{\arraystretch}{1.0}
\begin{tabular}{lcccc}
\toprule
{\textbf{Method}} & {\textbf{2WikiMultiHopQA}$\uparrow$} & {\textbf{MuSiQue}$\uparrow$} & {\textbf{Qasper}$\uparrow$} & {\textbf{Avg.} $\uparrow$} \\
\hline
\noalign{\vskip 1mm}
Llama2-7B-chat-4k & 29.50 & 6.50 & 17.00 & 17.67 \\
\midrule
\ + Ms-PoE~\citep{zhang2024middlelanguagemodelsuse} & 27.50 & 9.00 & \textbf{18.00} & 18.17 \\
\ + AB~\citep{chen-etal-2024-fortify}  & 31.00 & \textbf{11.00} & 16.50 & 19.50 \\
\ + MoICE~\citep{lin2024mixtureincontextexpertsenhance} & 30.00 & 10.00 & 15.50 & 18.50 \\
\ + \textit{PEAR} (Ours) & \textbf{35.00} & 8.50 & \textbf{18.00} & \textbf{20.50} \\
 \bottomrule
\end{tabular}
\end{table}

\begin{table}[t]
\centering
\caption{Practical inference time (in seconds) and GPU memory cost (in GB) per test sample for different methods. 
For a fair comparison, Flash-Attention~\citep{dao2023flashattention2} was not applied. 
The experiments were conducted on a single H800-80G GPU.}
\label{table:inference}
\vspace{5pt}
\resizebox{\linewidth}{!}{
\renewcommand{\arraystretch}{1.05}
\begin{tabular}{lcccc}
\toprule
\textbf{Method} & \textbf{2WikiMultiHopQA}$\downarrow$ & \textbf{MuSiQue}$\downarrow$ & \textbf{Qasper}$\downarrow$ & \textbf{Avg.}$\downarrow$\\ 
\hline
\noalign{\vskip 1mm}
Llama2-7B-chat-4k & 0.63/31.33 & 0.70/31.33 & 1.23/31.33 & 0.88/31.33 \\
\midrule
\ + Ms-PoE~\citep{zhang2024middlelanguagemodelsuse} & 0.95 (\textcolor{red}{+0.32}) / 39.21 (\textcolor{red}{+7.88}) & 1.11(\textcolor{red}{+0.41}) / 39.21 (\textcolor{red}{+7.88}) & 1.84 (\textcolor{red}{+0.61}) / 34.59 (\textcolor{red}{+3.26}) & 1.30 (\textcolor{red}{+0.42}) /37.67 (\textcolor{red}{+6.34}) \\
\ + AB~\citep{chen-etal-2024-fortify} & 2.50(\textcolor{red}{+1.87}) /66.19(\textcolor{red}{+34.86}) & 2.70(\textcolor{red}{+2.00}) / 66.19 (\textcolor{red}{+34.86}) & 5.67(\textcolor{red}{+4.44}) /66.34(\textcolor{red}{+35.01}) & 3.62(\textcolor{red}{+2.74})/66.24(\textcolor{red}{+34.91}) \\
\ + MoICE~\citep{lin2024mixtureincontextexpertsenhance} & 2.91(\textcolor{red}{+2.28}) / 79.13(\textcolor{red}{+47.80}) & 3.06(\textcolor{red}{+2.36}) / 79.12(\textcolor{red}{+47.79}) & 5.86(\textcolor{red}{+4.63}) / 79.10(\textcolor{red}{+47.77}) & 3.94(\textcolor{red}{+3.06}) / 79.12(\textcolor{red}{+47.79}) \\
\ + \textit{PEAR} (Ours) & 0.63 (\textcolor{blue}{\textbf{+0.00}}) /31.33 (\textcolor{blue}{\textbf{+0.00}}) & 0.70 (\textcolor{blue}{\textbf{+0.00}}) / 31.33 (\textcolor{blue}{\textbf{+0.00}}) & 1.23 (\textcolor{blue}{\textbf{+0.00}}) / 31.33 (\textcolor{blue}{\textbf{+0.00}}) & 0.88 (\textcolor{blue}{\textbf{+0.00}}) / 31.33 (\textcolor{blue}{\textbf{+0.00}}) \\
 \bottomrule
\end{tabular}}
\end{table}

\paragraph{\textbf{Detailed setups of proxy task}}

For head discovery, we constructed 200 task samples. 
In the case of the Llama and OPT models, we repeat the discovery process four times with varying values of \(n\): \(10, 15, 25, 50\). 
For the Baichuan model, the $n$ values are \(10, 20, 50, 80\). 
We found that each model has a group of heads with significantly large \(\Delta \pi\) values, leading us to select the \(K\) values based on the observed group sizes: 30, 22, and 21, respectively.
Table~\ref{tab:rag-head} presents the discovered RAG-suppression heads for the LLMs under study. 
Figure~\ref{fig:heads} illustrates \(\Delta \pi\) values when \(n=10\) for each model. 
Further detailed results can be found in Appendix~\ref{apx:head_res}. 

For the re-weighting coefficient learning stage, we constructed 500 task samples, setting \(n\) to 50 for all models.

\paragraph{\textbf{Hyperparameters for re-weighting coefficient learning}}

We employed the AdamW optimizer with a learning rate of 0.005 and parameters \((\beta_1, \beta_2) = (0.9, 0.999)\).
\(\tau\) are initialized at 1.0. 
Training was performed for a single epoch using BF16 precision on an A100-PCIE-40GB GPU. 

\paragraph{\textbf{Difference between learning and inference}}

We conducted internal loading training on the re-weighting coefficients $\tau^{(l,h)}$ during the training process. However, during inference, we externally re-weighting the output matrix weights of specific heads before loading the model. Therefore, our method is simple to train and has \textit{zero} inference overhead.

%


\subsection{Comparison with baselines on RAG tasks}

We compare \textit{PEAR} against various baselines on RAG tasks we constructed using three datasets: 2WikiMultihopQA~\citep{xanh2020_2wikimultihop}, MuSiQue~\citep{trivedi-etal-2022-musique}, and Qasper~\citep{dasigi2021datasetinformationseekingquestionsanswers}. The first two datasets require the model to answer questions based on multiple documents, while the third focuses on questions related to NLP research papers, formulated and answered by NLP researchers. We truncate the context to 4,000 tokens for the first two datasets; the third dataset has an average context length of 3,619 tokens.

Our experiments are conducted with Llama2-7B-chat-4k, as the baselines are tailored specifically for RoPE. 
We evaluate the models' performance using exact match scores, with results reported in Table~\ref{table:lb}. 
Notably, our method achieves the highest average improvement across all three tasks.
Although \textit{PEAR} does not achieve the top performance on the MuSiQue task, it outperforms the original model by a large margin. 

Additionally, we present inference time and memory costs for these datasets in Table~\ref{table:inference}. 
\textit{PEAR} does not increase GPU memory usage and inference time costs.
This makes it significantly more efficient than other enhancement methods.

These experiments underscore the effectiveness and efficiency of \textit{PEAR} in enhancing LLMs for RAG tasks.

\subsection{Applicability to LLMs using various position embeddings}

In this section, we demonstrate the applicability of \textit{PEAR} to LLMs utilizing different position embeddings. 
We conduct a multi-document question-answering (MDQA) experiment based on data from~\citep{liu2023lost}, which leverage a subset of NaturalQuestions-Open ~\citep{lee2019latent, kwiatkowski2019natural}, consisting of 2,655 queries. Each query is paired with a context consisting of 10 documents with an average of 1,722 tokens.
Following~\citep{liu2023lost,chen-etal-2024-fortify}, we position the gold document (i.e., the document contains the ground truth answer) at various contextual positions to evaluate the robustness of a context-awareness enhancement method. 
In our experiments, we set the maximum document count to 10 and assess the question-answering accuracy when the gold document is placed as the 1st, 3rd, 5th, 7th, and 10th document, respectively. 
Since baseline methods are not compatible with the OPT and Baichuan models, we compare \textit{PEAR} only with the original models. 
The results are presented in Table~\ref{table:nih}.

\begin{table}[ht]
\centering
\caption{Experimental results on the MDQA task show that \textit{PEAR} achieves the highest accuracy in 24 out of 25 comparisons across three LLMs, demonstrating its broad applicability to various position embeddings and its robustness in enhancing awareness to different contextual positions.}
\label{table:nih}
\vspace{5pt}
\resizebox{\linewidth}{!}{
\begin{tabular}{@{}clcccccc@{}}
\toprule
\multirow{2}{*}{\textbf{Position Embedding}} & \multirow{2}{*}{\textbf{Method}} & \multicolumn{5}{c}{\textbf{Gold Document Position}} & \multirow{2}{*}{\textbf{Avg.}} \\
\cmidrule(lr){3-7}
 & & \textbf{1} & \textbf{3} & \textbf{5} & \textbf{7} & \textbf{10} \\
\midrule
\multirow{5}{*}{RoPE} & Llama2-7B-chat-4k & 64.14 & 65.95 & 64.97 & 62.67 & 67.53 & 65.05 \\
 & \ + Ms-PoE~\citep{zhang2024middlelanguagemodelsuse} & 66.06 & 64.29 & 63.99 & 62.22 & 64.75 & 64.34 \\
 & \ + AB~\citep{chen-etal-2024-fortify} & \textbf{66.36} & 66.14 & {65.25} & {63.20} & 64.93 & {65.18} \\
 & \ + MoICE~\citep{lin2024mixtureincontextexpertsenhance} & 65.50 & 66.33 & 65.61 & 64.11 & {65.84} & 65.48 \\ 
 & \ + \textit{PEAR} (Ours) & 62.71 & \textbf{67.01} & \textbf{68.32} & \textbf{66.44} & \textbf{69.57} & \textbf{66.81}\\
\midrule
\multirow{2}{*}{Learnable Embeddings} & OPT-6.7B-2k & 19.07 & 15.45 & 17.03 & 16.54 & 22.61 & 18.14 \\
  & \ + \textit{PEAR} (Ours) & \textbf{20.23} & \textbf{17.18} & \textbf{17.60} & \textbf{17.22} & \textbf{22.87} & \textbf{19.02}\\
\midrule
\multirow{2}{*}{Alibi} & Baichuan-13B-chat-4k & 12.28 & 13.45 & 11.98 & 11.04 & 12.96 & 12.34 \\
  & \ + \textit{PEAR} (Ours) & \textbf{14.16} & \textbf{14.84} & \textbf{13.67} & \textbf{12.77} & \textbf{13.94} & \textbf{13.88}
 \\\bottomrule
\end{tabular}}
\end{table}

\subsection{\textit{PEAR} does not diminish knowledge capabilities in LLMs}
\label{sec:safe}

Previous research \citep{geva-etal-2023-dissecting, lv2024interpretingkeymechanismsfactual} has shown that certain attention heads store or play a crucial role in eliciting parametric knowledge. 
This raises the question of whether \textit{PEAR} enhances context awareness of LLMs at the expense of their ability to utilize this parametric knowledge. 

To investigate this, we evaluated a \textit{PEAR}-enhanced Llama2-7B-chat model using the MMLU benchmark \citep{mmlu}, and the results are presented in Table~\ref{table:mmlu}. 
The performance of the enhanced Llama2-7B-chat and the original Llama2-7B-chat did not show a significant difference. 
Consequently, we argue that \textit{PEAR}, through its effective head discovery and re-weighting learning approaches, does not compromise the knowledge capabilities of LLMs.

\begin{table}[!h]
\centering
\caption{Results on the MMLU benchmark showing that \textit{PEAR} does not enhance LLMs' context capacities at the expense of knowledge ability.}
\label{table:mmlu}
\vspace{5pt}
\resizebox{0.8\linewidth}{!}{
\begin{tabular}{@{}l|ccccc@{}}
\toprule
\textbf{Model} & \textbf{Humanities} & \textbf{Social Science} & \textbf{STEM} & \textbf{Other}  & \textbf{Avg.}\\
\midrule
Llama2-7B-chat-4k & 42.55 & 52.29 & 37.14 & 52.47 & 45.81 \\
\ + \textit{PEAR} (Ours) & 42.06 & 52.03 & 36.61 & 52.19 & 45.41 \\
\bottomrule
\end{tabular}}
\end{table}

\begin{table}[h]
\centering
\caption{The experiment results on the question answering task with ablation settings, which show that our control over the number of suppression heads is effective.}
\label{table:ablationlongbench}
\vspace{5pt}
\renewcommand{\arraystretch}{1.05} 
\begin{tabular}{lcccc}
\toprule
{\textbf{Method}} & {\textbf{2WikiMultiHopQA}} & {\textbf{MuSiQue}} & {\textbf{Qasper}} & {\textbf{Avg.}} \\
\hline
\noalign{\vskip 1mm}
Llama2-7B-chat-4k & 29.50 & 6.50 & 17.00 & 17.67 \\
\midrule
\ + \textit{PEAR} ($K$=10) & 33.00 & 8.00 & 16.00 & 19.00 \\
\ + \textit{PEAR} ($K$=20) & 33.50 & 8.50 & 16.50 & 19.50 \\
\ + \textit{PEAR} ($K$=30) & \textbf{35.00} & \textbf{8.50} & \textbf{18.00} & \textbf{20.50} \\
\ + \textit{PEAR} ($K$=40) & 32.50 & 8.00 & 17.00 & 19.17 \\
 \bottomrule
\end{tabular}
\end{table}

\begin{table}[!ht]
\centering
\caption{The experiment results on the MDQA task with ablation settings, which show that our control over the number of suppression heads is effective.}
\label{table:ablationlmdqa}
\begin{tabular}{@{}l|cccccc@{}}
\toprule
\multirow{2}{*}{\textbf{Method}} & \multicolumn{5}{c}{\textbf{Gold Document Position}} & \multirow{2}{*}{\textbf{Avg.}} \\
\cmidrule(lr){2-6}
 & \textbf{1} & \textbf{3} & \textbf{5} & \textbf{7} & \textbf{10} \\
\midrule
Llama2-7B-chat-4k & \textbf{64.14} & 65.95 & 64.97 & 62.67 & 67.53 & 65.05 \\
\midrule
\ + \textit{PEAR} ($K$=10) & 63.43 & 66.26 & 66.82 & 64.67 & 67.76 & 65.79 \\ 
\ + \textit{PEAR} ($K$=20) & 63.88 & 66.29 & 66.44 & 65.65 & 68.51 & 66.15 \\
\ + \textit{PEAR} ($K$=30) & 62.71 & \textbf{67.01} & \textbf{68.32} & 66.44 & \textbf{69.57} & \textbf{66.81}\\
\ + \textit{PEAR} ($K$=40) & 62.90 & 66.00 & 67.16 & \textbf{66.59} & 68.40 & 66.21 \\
\bottomrule
\end{tabular}
\end{table}

\subsection{Analysis: The effect of \texorpdfstring{$K$}{}}

While we have demonstrated the effectiveness of \textit{PEAR} from various angles, a key point for discussion is the role of \( K \), representing the number of heads to re-weight.

Using Llama2-7B-chat as a case study, we vary \( K \) and observe its impact on \textit{PEAR}'s performance. 
Table~\ref{table:ablationlongbench} presents results on RAG tasks, while Table~\ref{table:ablationlmdqa} details analysis for MDQA tasks. 
The findings indicate that \textit{PEAR} performs optimally when \( K \) matches the inherent threshold of the model (i.e., $K=30$ for Llama2-7B-chat), i.e., the number of heads with a significantly higher \(\Delta\pi\) than others.

Re-weighting fewer heads fails to fully alleviate the suppression from RAG-suppression heads, while exceeding this optimal number can harm the performance of non-RAG-suppression heads, ultimately diminishing overall effectiveness.

\subsection{Analysis: The Value of \texorpdfstring{$\tau$}{}}
\label{sec:tau}
\begin{wrapfigure}{r}{0.55\linewidth}
    \vspace{-14mm}
    \centering
    \includegraphics[width=\linewidth]{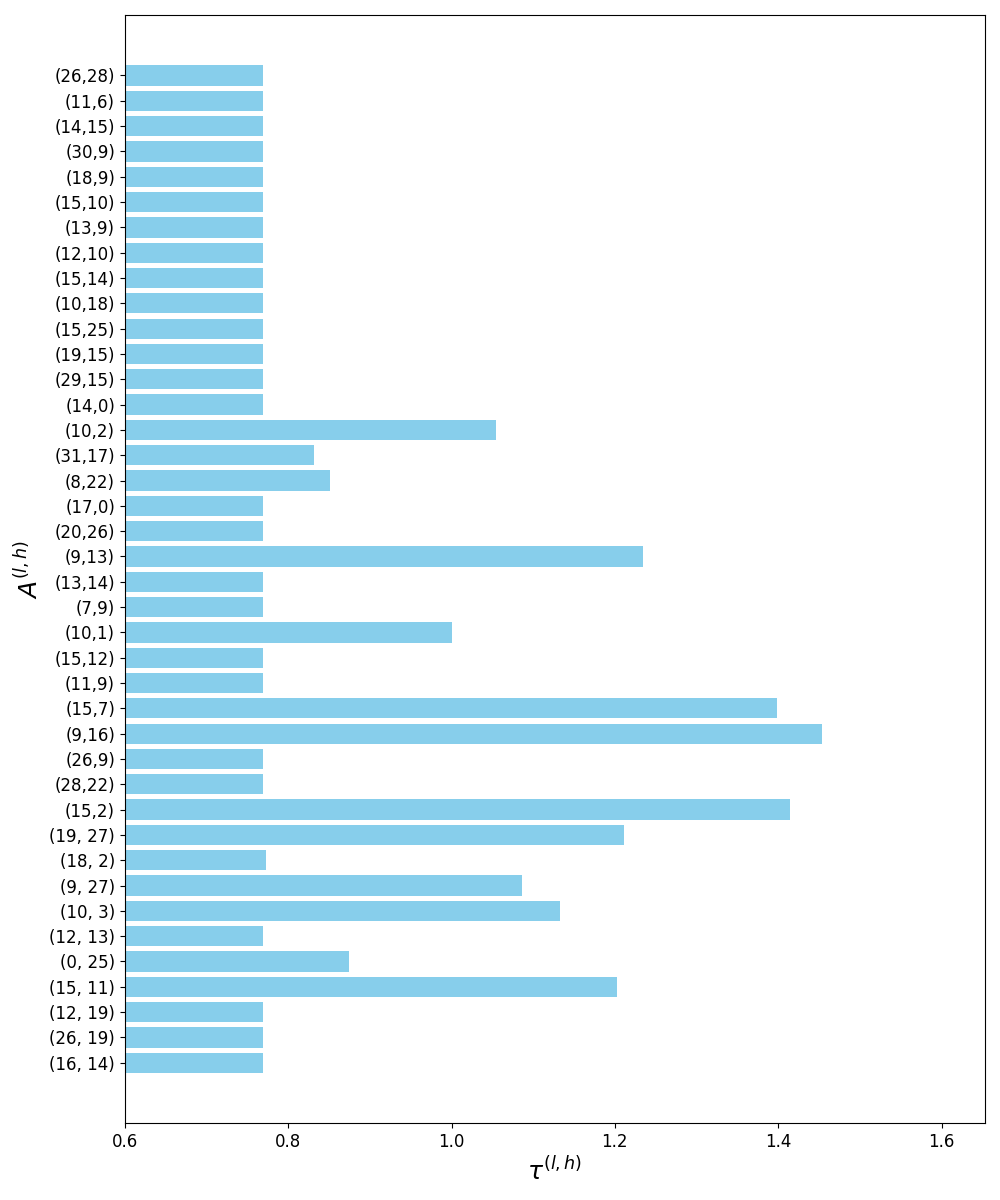}
    \caption{The learned coefficients of \textit{PEAR} (on Llama2-7B-chat and $K=40$).}
    \label{fig:params}
    \vspace{-6mm}
\end{wrapfigure}
Using Llama2-7B-chat as an example, we present the learned coefficients of PEAR in Figure~\ref{fig:params}, with heads ranked by their $\Delta\pi$ scores.

Intuitively, most heads are optimized to values less than one, which reduces their relative weight compared to other heads within the same layer when multi-head outputs are aggregated. 
Due to BF16 training precision, many $\tau$s are optimized to the same value.
However, using FP32 precision for training did not significantly impact the results.
Notably, $\tau$s for 9 heads, which have relatively low $\Delta\pi$, are greater than one. 
We do not attribute this to the precision of the discovery process, as constraining the re-weighting coefficients to be less than one led to suboptimal performance. Thus, a plausible explanation is that RAG suppression is a complex, cooperative effect involving multiple heads, each with distinct working mechanisms, as discussed in Section~\ref{head_detection}.

\section{Conclusion}

In this paper, we introduce \textit{PEAR}, a position-embedding-agnostic method designed to enhance the performance of LLMs on RAG tasks with zero inference overhead. 
Our method not only outperforms competitive baselines in both effectiveness and efficiency but also demonstrates broad applicability across various LLMs. 
We also presented that \textit{PEAR} improves context awareness in LLMs without compromising their inherent knowledge capabilities. 
These benefits make \textit{PEAR} a promising approach for a wide range of applications that require robust context abilities, such as in-context learning and strict instruction following, which we leave for future research.


\newpage
\bibliography{iclr2025_conference}
\bibliographystyle{iclr2025_conference}

\appendix
\section{Head discovery results}
\label{apx:head_res}

\begin{figure}[h]
\vspace{-2mm}
    \centering
    \includegraphics[width=0.98\linewidth]{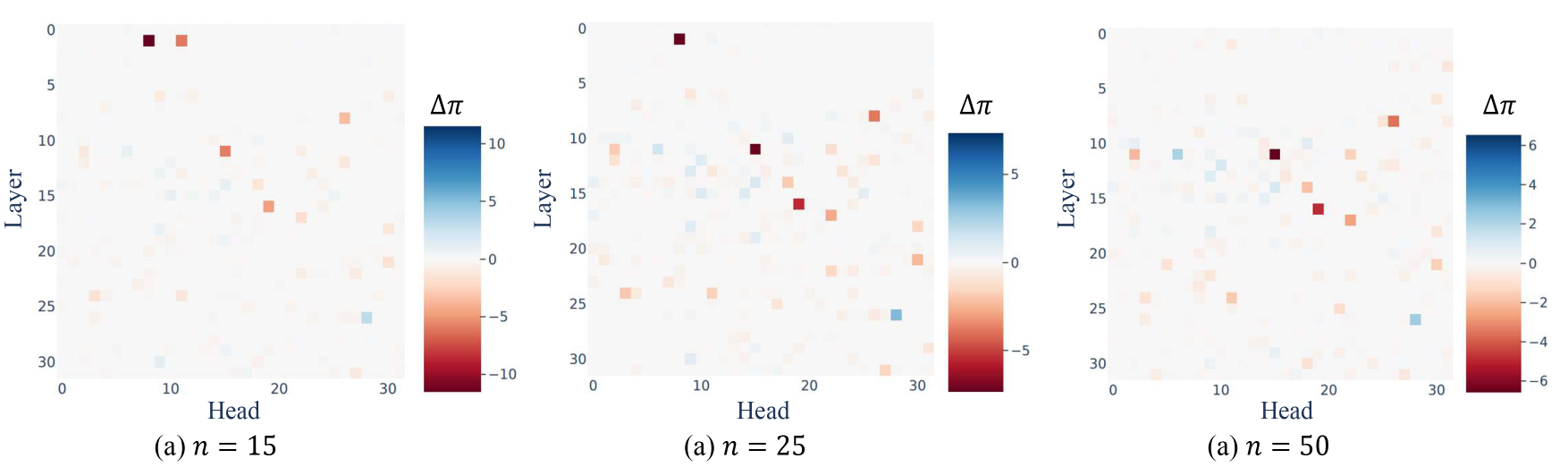}
    \caption{Heatmaps of $\Delta\pi$ scores for each head of llama2-7B-chat ($n=15, n=25, n=50$).}
    \label{fig:llama_heads}
    \vspace{-2mm}
\end{figure}

\begin{figure}[h]
\vspace{-2mm}
    \centering
    \includegraphics[width=0.98\linewidth]{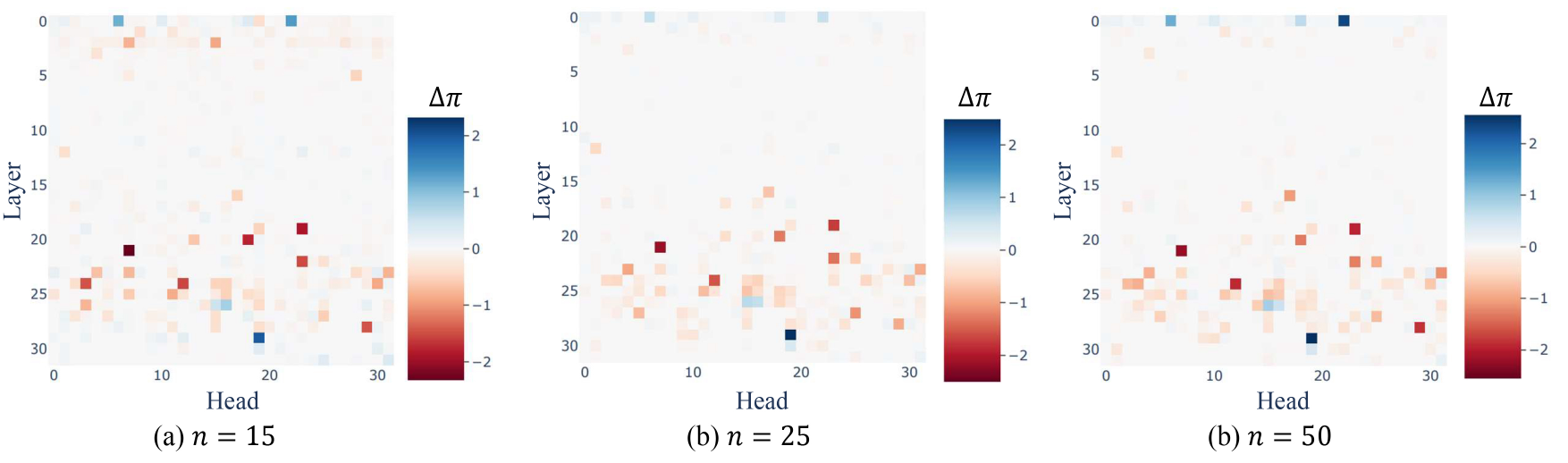}
    \caption{Heatmaps of $\Delta\pi$ scores for each head of OPT-6.7B ($n=15, n=25, n=50$).}
    \label{fig:opt_heads}
    \vspace{-2mm}
\end{figure}

\begin{figure}[h]
\vspace{-2mm}
    \centering
    \includegraphics[width=0.98\linewidth]{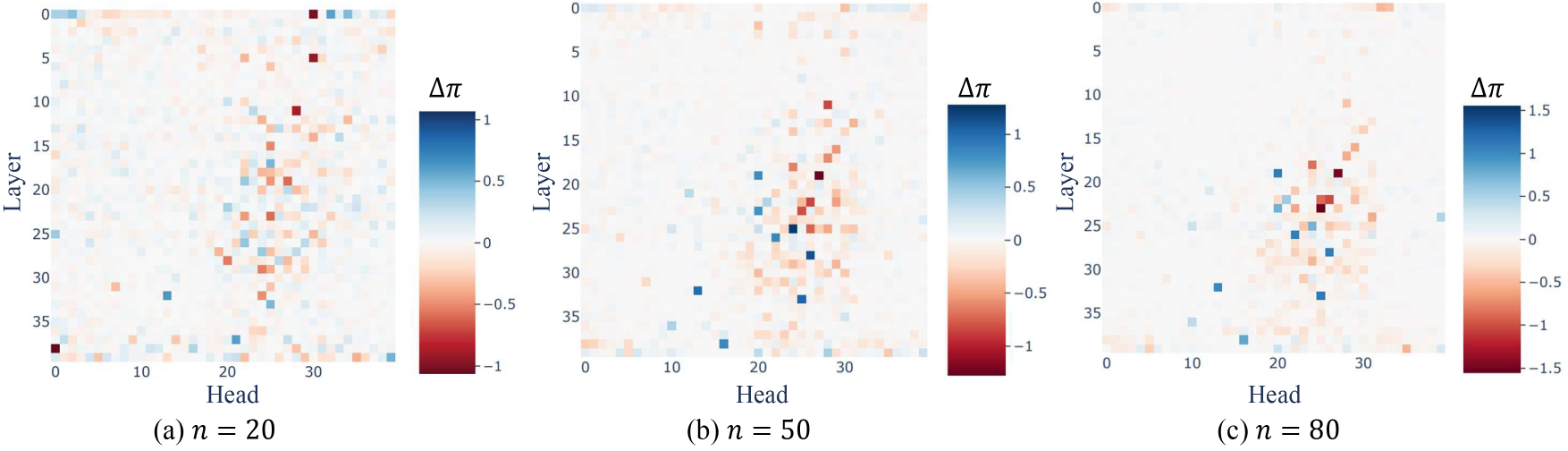}
    \caption{Heatmaps of $\Delta\pi$ scores for each head of Baichuan-13B-chat ($n=20, n=50, n=80$).}
    \label{fig:baichuan_heads}
    \vspace{-2mm}
\end{figure}

\end{document}